\documentclass[10pt,twocolumn,letterpaper]{article}

\usepackage{cvpr}      
\usepackage[dvipsnames]{xcolor}

\usepackage{graphicx}
\usepackage{amsmath}
\usepackage{bm}
\usepackage{amssymb}
\usepackage{booktabs}
\usepackage{booktabs}
\usepackage{threeparttable}
\usepackage{multirow}
\usepackage{bbding}
\toggletrue{cvprfinal}

\definecolor{cvprblue}{rgb}{0.21,0.49,0.74}
\usepackage[pagebackref,breaklinks,colorlinks]{hyperref}

\usepackage[capitalize]{cleveref}
\crefname{section}{Sec.}{Secs.}
\Crefname{section}{Section}{Sections}
\Crefname{table}{Table}{Tables}
\crefname{table}{Tab.}{Tabs.}

\def\paperID{392} 
\def\confName{CVPR}
\def\confYear{2024}
\title{Distilling Semantic Priors from SAM to Efficient Image Restoration Models}

\author{
Quan Zhang$^{1,3,}$\thanks{Both authors contributed equally to this research, which was done during Quan Zhang and Xiaoyu Liu's internship at Huawei Noah’s Ark Lab.}\quad
Xiaoyu Liu$^{2,3,}$\footnotemark[1]\quad
Wei Li$^{3}$ \quad
Hanting Chen$^{3}$ \quad
Junchao Liu$^{3}$ \quad
Jie Hu$^{3}$ \quad\\
Zhiwei Xiong$^{2}$ \quad
Chun Yuan$^{1,}$ \thanks{Corresponding authors: 
\begin{tabular}[t]{@{}l@{}} 
    yunhe.wang@huawei.com
\end{tabular}} \quad
Yunhe Wang$^{3,}$ \footnotemark[2] \quad
\and
$^1$Tsinghua Shenzhen International Graduate School \\
$^2$University of Science and Technology of China \quad
$^3$Huawei Noah’s Ark Lab}

\begin{document}




\maketitle

\begin{abstract}
In image restoration (IR), leveraging semantic priors from segmentation models has been a common approach to improve performance.  The recent segment anything model (SAM) has emerged as a powerful tool for extracting advanced semantic priors to enhance IR tasks.  However, the computational cost of SAM is prohibitive for IR, compared to existing smaller IR models. The incorporation of SAM for extracting semantic priors considerably hampers the model inference efficiency. To address this issue, we propose a general framework to distill SAM's semantic knowledge to boost exiting IR models without interfering with their inference process. Specifically, our proposed framework consists of the semantic priors fusion (SPF) scheme and the semantic priors distillation (SPD) scheme. SPF fuses two kinds of information between the restored image predicted by the original IR model and the semantic mask predicted by SAM for the refined restored image. SPD leverages a self-distillation manner to distill the fused semantic priors to boost the performance of original IR models. Additionally, we design a semantic-guided relation (SGR) module for SPD, which ensures semantic feature representation space consistency to fully distill the priors. We demonstrate the effectiveness of our framework across multiple IR models and tasks, including deraining, deblurring, and denoising.
\end{abstract}

\begin{figure}[!t]
  \centering
  \includegraphics[width= 0.47\textwidth]{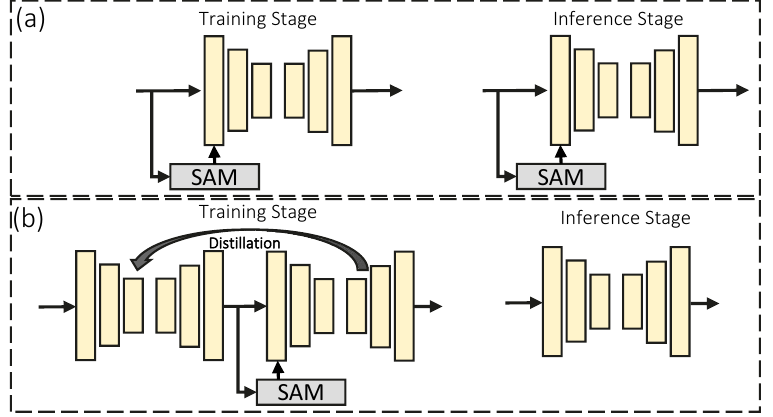}
  \vspace{-0.3cm}
  \caption{Comparison of training and inference pipelines between different manners of exploiting semantic priors from SAM.
  (a) Existing methods require the use of SAM at both the training and inference stages. (b) Our method only uses SAM at the training stage and preserves the same inference efficiency as the original image restoration model at the inference stage.} 
  \label{teaser}
  \vspace{-0.6cm}
\end{figure}

\section{Introduction}
\label{sec:intro}
Image restoration (IR)~\cite{kopf2008deep,geman1984stochastic,he2010single} is an essential computer vision task that reconstructs high-quality (HQ) images from degraded low-quality (LQ) inputs. These inputs are impaired by distortions like noise~\cite{zhang2021plug,kulikov2023sinddm}, blurring~\cite{zhang2019deep,li2022learning,dong2023enhanced}, and rain drops~\cite{wang2020model,guo2021efficientderain}. To address this, IR methods incorporate explicit image priors and models of the distortion process. These constraints help narrow the solution space for feasible reconstruction. With the advent of deep learning~\cite{zhang2020residual,zhang2021plug,zamir2022restormer,dong2023dfvsr}, data-driven techniques have achieved superior performance in IR tasks by learning strong statistical regularities. 
Previous studies ~\cite{wu2023learning,wei2022sginet,wang2018recovering,li2022close} have demonstrated the potential benefits of utilizing semantic priors obtained from segmentation for image restoration tasks. These priors contain valuable information about the texture and color characteristics of individual objects within an image. By incorporating these priors, a deeper understanding of the image content can be achieved, providing explicit instructions that guide the restoration process. This integration of semantic priors enhances the restoration performance by leveraging the rich knowledge encoded within the segmentation results.

In recent years, the emergence of the Segment Anything Model (SAM)~\cite{kirillov2023segment} has had a profound impact on various computer vision tasks~\cite{cheng2023tracking,xie2023edit,yang2023track,gong20233dsam}, including image restoration~\cite{li2023sam,xiao2023dive,jin2023let,lu2023can}. By scaling up datasets and model capacity, this large foundation model exhibits capabilities beyond surface-level labeling and provides useful cues unavailable to mainstream networks. Specifically, SAM can extract advanced semantic priors through a holistic understanding of image content. However, the computational cost associated with SAM poses a significant challenge when applied to IR, particularly when compared to existing smaller IR models. Incorporating SAM for extracting semantic priors during the inference stage considerably hampers the overall efficiency of the model. As shown in Fig.~\ref{teaser} (a), Existing methods require the use of SAM at both the training and inference stages. The SAM module is employed during training to learn spatial attention for capturing semantic priors. However, during inference, the SAM module is also utilized, which can potentially introduce additional computational overhead.

To address this issue, we propose a general framework that distills SAM's semantic knowledge to enhance existing IR models without interfering with their inference process. As shown in Fig.~\ref{teaser}(b), the objective of our framework is to leverage the benefits of SAM's semantic priors while mitigating the computational burden and preserving the efficiency of the IR models. Our proposed framework consists of two key schemes: the semantic priors Fusion (SPF) scheme and the semantic priors Distillation (SPD) scheme with a semantic-guided relation (SGR) module. The SPF scheme focuses on fusing information from two sources: the restored image predicted by the original IR model and the semantic mask predicted by SAM. The SPD scheme aims to distill the fused semantic priors to boost the performance of the original IR model. 
In the SPF scheme, we combine these two sources as inputs of the  IR network cascaded behind the original IR network, to refine the restored image and improve its quality. This fusion process enables the incorporation of SAM's semantic knowledge into the IR model without sacrificing efficiency.
In the SPD scheme, we leverage the knowledge distillation manner~\cite{pham2022revisiting,liu2024graph,liu2022efficient} to distill the fused semantic priors obtained through the SPF scheme by enforcing the consistency between the original restored image and the refined original restored image. This scheme aims to enhance the performance of the original IR model by transferring and consolidating the valuable insights extracted by SAM. Additionally, we design a semantic-guided relation  (SGR) module for SPD, which ensures consistency in the semantic feature representation space. This further enhances the distilled priors and promotes their effectiveness in improving the IR model's performance. Finally, we only utilize the distilled IR model to restore the degraded LQ inputs without segmentation masks from SAM.

To validate the effectiveness of our proposed framework, we conduct extensive experiments across multiple IR models and tasks, including deraining, deblurring, and denoising. The results demonstrate the potential of our approach in leveraging SAM's semantic knowledge to enhance the performance of existing IR models while addressing the computational challenges associated with SAM integration.

Overall, our contributions can be summarized as follows:
\begin{itemize}
\item We propose a general framework to distill semantic knowledge from SAM to boost existing IR models without interfering with their inference process.
\item We propose an SPF scheme to fuse information between restored images from the IR model and semantic masks from SAM to refine the restoration.
\item We propose an SPD scheme that uses a self-distillation manner with a designed semantic-guided relation module to transfer semantic priors from SPF into the original IR models.
\item The effectiveness of our framework is demonstrated through experiments on various IR models and tasks.
\end{itemize}

\begin{figure*}[!t]
  \centering
  \includegraphics[width=\textwidth]{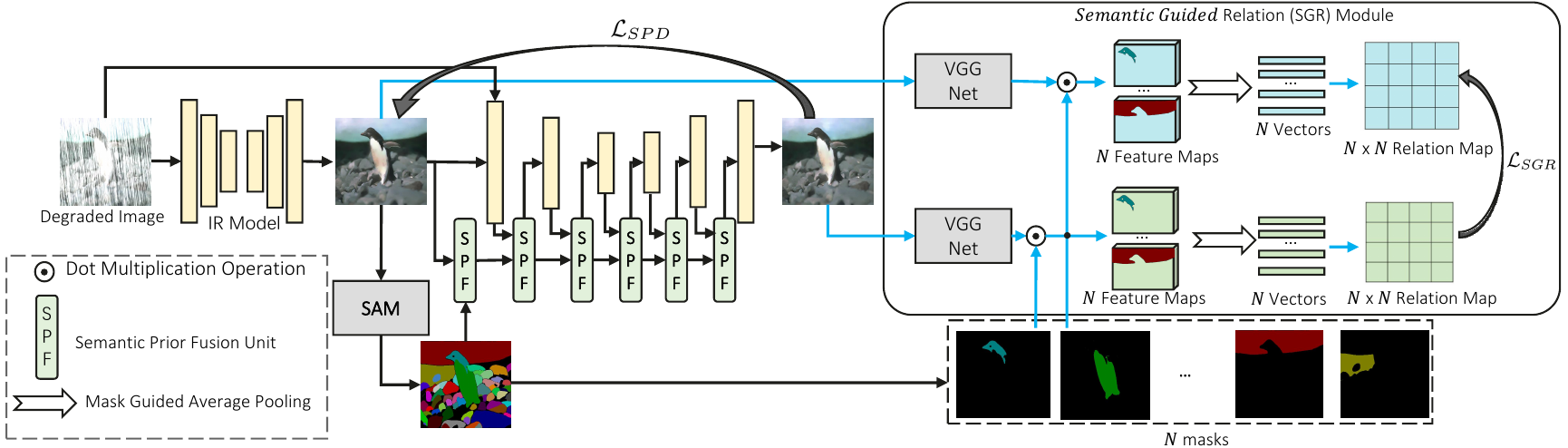}
  \vspace{-0.7cm}
  \caption{The workflow of our proposed framework to distill semantic knowledge from SAM to boost existing IR models without interfering with their inference process.} 
  \vspace{-0.5cm}
  \label{workflow}
\end{figure*}

\section{Related Work}

\noindent\textbf{Semantic Priors for Image Restoration.}
Existing methods can deal with the degraded images by low-level and high-level vision interaction, which are categorized into two types~\cite{wu2023learning}: loss-level methods and feature-level methods.
Loss-level methods~\cite{liu2017image,zheng2022semantic,wang2020dual} focus on incorporating semantic priors by utilizing semantic-aware losses as additional objective functions during the training process of original vision tasks. However, these methods exploit semantic priors in an implicit manner, lacking sufficient interaction between semantic priors and IR tasks.
Feature-level methods~\cite{ren2018deep,shen2020exploiting,li2022close,wang2018recovering} integrate semantic priors into the feature representation space by extracting intermediate features from semantic segmentation networks and combining them with image features. While these methods explicitly exploit semantic priors to significantly improve the performance of IR tasks, these methods modify the inference way of original IR models and still rely on the input of semantic inputs during the inference stage.
Although some recent works have attempted to combine both loss-level and feature-level methods~\cite{wu2023learning}, existing methods still do not adequately exploit the semantic priors of the Segment Anything Model (SAM) to achieve sufficient interaction and inference efficiency. Our framework ensures effective interaction between semantic priors and IR tasks without interfering with the inference process of the IR models.

\noindent\textbf{SAM VS. Other Segmentation Models.}
SAM is a foundation model for the image segmentation task with zero-shot generalization capacity, which can be used to solve a range of downstream segmentation problems on new data distributions using prompt engineering. 
Although existing segmentation models have achieved excellent performance in various segmentation tasks such as semantic segmentation~\cite{jain2023semask,long2015fully,zhao2017pyramid}, instance segmentation~\cite{he2017mask,fang2021instances,liu2023learning}, panoptic segmentation~\cite{kirillov2019panoptic,xiong2019upsnet}, and unified segmentation~\cite{li2023mask}, these models heavily rely on annotated segmentation masks for training. However, in many image restoration tasks, such segmentation annotations are not available. This is where SAM stands out with its zero-shot generalization capability. SAM can effectively handle image restoration tasks without the need for specific segmentation annotations.
In addition, SAM provides more fine-grained semantic information compared to instance and category labels through segmentation at different granularity. This finer level of semantic information is crucial for image restoration tasks, as it enables the utilization of richer semantic priors in the restoration process.

\section{Method}
In this section, we introduce our proposed framework, as illustrated in Fig~\ref{workflow}. This framework comprises two essential schemes: the semantic priors Fusion (SPF) scheme (detailed in Sec.~\ref{spf} ) and the semantic priors Distillation (SPD) scheme with a semantic-guided relation (SGR) module ((detailed in Sec.~\ref{spd})). The SPF scheme focuses on fusing information from two distinct sources: the restored image predicted by the original IR model and the semantic mask predicted by SAM. On the other hand, the SPD scheme aims to distill the fused semantic priors to enhance the performance of the original IR model.

\subsection{Problem Analysis and Definition}
Semantic priors offer invaluable guidance for restoring colors, contrasts, and texture consistency in degraded images. As a large-scale foundation model, SAM contains extensive parameters trained on diverse distributions of image data. This broad exposure equips SAMs with rich semantic knowledge to inform restoration processes. By leveraging SAMs' understanding of semantic concepts and contexts, we can provide restoration models with informative cues to improve fidelity.
Moreover, the scale and generalizability of the SAM allow for capturing higher-order semantics beyond the scope of restoration datasets. These holistic scene-level cues can further boost coherence in restored images. 





Given a degraded low-quality image $I_{LQ}\in \mathbb{R}^{3 \times H \times W}$, we can obtain its segmentation mask by utilizing SAM's automatic mode:
\begin{equation}
\begin{aligned}
M_{LQ} = f_{SAM}(I_{LQ}),
\end{aligned}
\end{equation}
where, $M_{LQ}$ represents the semantic priors extracted from the degraded image $I_{LQ}$ and is the segmentation masks of each objects, and $f_{SAM}$ refers to the SAM model.

The objective of existing solutions~\cite{li2023sam,xiao2023dive,jin2023let} is to utilize both the semantic priors $M_{LQ}$ and the degraded input image $I_{LQ}$ to restore a high-quality image $I_{HQ}$, represented as $(M_{LQ},I_{LQ})\rightarrow I_{HQ}$. In our framework, we propose two distinct designs to overcome two drawbacks in the conventional pipeline:

1) Directly combining the semantic priors $M_{LQ}$ from SAM and the input image $I_{LQ}$ at the feature level is impractical since it hampers the inference efficiency of the original IR model and requires the input of SAM's prior during the inference stage. To address this, we cascade two IR models, $f^{IR1}$ and $f^{IR2}$, together. The second IR model, $f^{IR2}$, is responsible for fusing the semantic priors, and then the fused model's capabilities are distilled to the first IR model, $f^{IR1}$. This allows us to utilize $f^{IR1}$ during the inference process, thereby maintaining the efficiency of the original IR model.

2) Since the semantic priors $M_{LQ}$ is extracted from the severely degraded image $I_{LQ}$, it may contain segmentation errors. To address this, we propose to extract the semantic priors from the restored image obtained from $f^{IR1}$:
\begin{equation}
\begin{aligned}
I_{HQ}^1 = f_{IR1}(I_{LQ}),\
M = f_{SAM}(I_{HQ}^1),
\end{aligned}
\end{equation}
where, $I_{HQ}^1$ represents the output of $f^{IR1}$, and $M\in \mathbb{R}^{N \times H \times W}$ is the segmentation with $N$ mask channels.

\begin{figure}[!t]
  \centering
  \includegraphics[width= 0.47\textwidth]{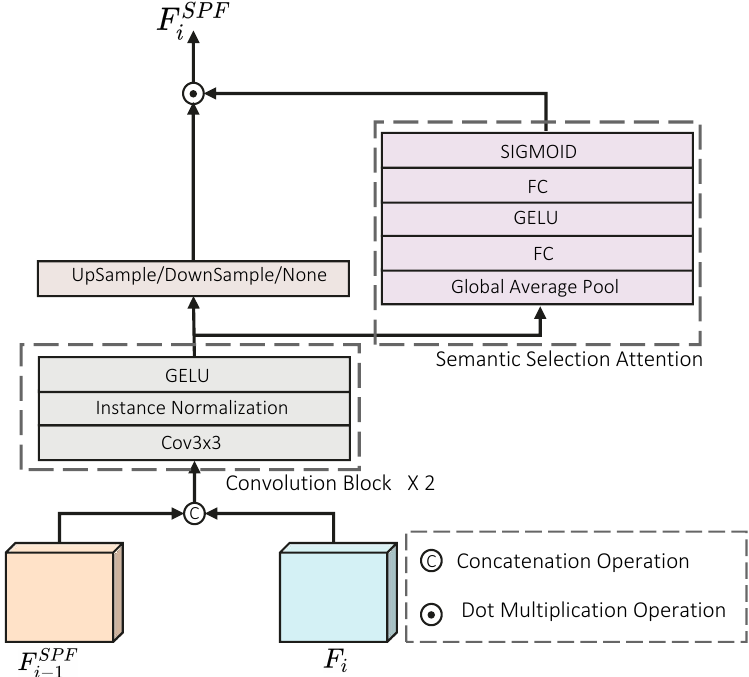}
\vspace{-0.3cm}
  \caption{Architecture of the Semantic Prior Fusion (SPF) unit.} 
  \label{SPF}
  \vspace{-0.4cm}
\end{figure}

\subsection{Semantic Priors Fusion}
\label{spf}
The SPF scheme is used to fuse the semantic priors $M$ and preliminary restored image $I_{HQ}^1$ from the IR model $f^{IR1}$ into the second IR model $f^{IR2}$ for obtaining the enhanced restored image $I_{HQ}^2$. 
To provide a concrete example of how the SPF (Semantic Prior Fusion) scheme works, let's consider the feature maps $F_i$ from the $i^{th}$ building block of the image restoration model $f^{IR2}$. The SPF scheme consists of multiple SPF units, where the number of SPF units matches the number of building blocks in $f^{IR2}$. Each SPF unit combines the semantic prior $M$ extracted from the restored image and the corresponding feature maps $F_i$.

Without loss of generality, this formulation exemplifies how SPF incorporates semantic guidance into intermediate feature representations within the IR model $f^{IR2}$:
\begin{equation}
\begin{aligned}
&F_{i+1} = f(F_i^{SPF} ),\\
&\begin{cases}
  F_i^{SPF} = f_i^{SPF}(\left [ F_{i-1}^{SPF}, F_{i} \right ])  \text{ if } i>1 \\
  F_i^{SPF} = f_i^{SPF}(\left [ I_{HQ}^1, M \right ])    \text{ if } i=1
\end{cases},
\end{aligned}
\end{equation}
where $f(\cdot )$ represents the next building block to generate the next feature maps, $f_i^{SPF}$ represents the $i^{th}$ SPF unit in the designed SPF scheme to fuse the feature map $F_i$ and the semantic prior $M$ from SAM. $\left [ \cdot \right ]$ represents the concatenation operation.


The architecture of the proposed SPF unit $f_i^{SPF}$ is detailed in Fig.~\ref{SPF}. We first concatenate the input feature maps $F_{i-1}^{SPF}$ and $F_{i}$. This combined input is fed into two convolutional blocks to extract an intermediate fused representation.
To align with the designs of different IR network building blocks, we introduce resize operations to match dimensions as needed. Additionally, a Semantic Selection Attention module is proposed to focus the SPF fusion on effective semantic priors. It implements a dot product to selectively highlight salient semantics. 

By repeating this SPF unit across network stages, semantic guidance is systematically integrated into the IR model.

\subsection{Semantic Priors Distillation}
\label{spd}


During the initial training stage, the output of $f^{IR2}$ is the high-quality restored image $I_{HQ}^2$, which aims to incorporate semantic priors and has better image quality compared to the restored image $I_{HQ}^1$ from the output of $f^{IR1}$. To transfer the capabilities of $f^{IR2}$ to $f^{IR1}$ and improve the performance of $f^{IR1}$ to match or approach the performance of $f^{IR2}$, we propose a semantic priors distillation scheme. This scheme facilitates the convergence of both networks during training.

In the semantic priors distillation scheme, we leverage the high-quality restored image $I_{HQ}^2$ obtained from $f^{IR2}$ as a teacher signal to guide the training of $f^{IR1}$. The key points of the scheme are as follows:

Firstly, we distill the semantic priors from $f^{IR2}$ to $f^{IR1}$ by minimizing the smooth L-1 loss between $I_{HQ}^1$ and $I_{HQ}^2$. In image restoration tasks, there may be cases where the restored images contain artifacts. By using the smooth L1 loss, we can reduce the influence of these artifacts and provide more robust training:
\begin{equation}
\small
\begin{aligned}
\mathcal{L}_{SPD} =
\begin{cases}
  \left \| I_{HQ}^1, - I_{HQ}^2 \right \|_1-0.5 & \text{ if } \left \| I_{HQ}^1, - I_{HQ}^2 \right \|_1 >1, \\
0.5 \times \left \| I_{HQ}^1, - I_{HQ}^2 \right \|_1^2 & \text{ if } \left \| I_{HQ}^1, - I_{HQ}^2 \right \|_1<1,
\end{cases}
\end{aligned}
\end{equation}
where $\left \| \cdot \right \|_1$ represents the L1 loss. Through minimizing the distillation loss, we encourage the semantic priors extracted by $f^{IR1}$ to converge to the semantic priors from $f^{IR2}$ as the two networks are jointly optimized.

Secondly, we introduce a semantic-guided relation (SGR) module to facilitate the transfer of semantic priors from $f^{IR2}$ to $f^{IR1}$ while ensuring consistency between $I_{HQ}^1$ and $I_{HQ}^2$ in the semantic feature representation space. To achieve this, we utilize a pre-trained VGG model to extract semantic-aware feature maps $F_{VGG}^1 \in \mathbb{R}^{512 \times \frac{H}{8} \times \frac{W}{8} }$ and $F_{VGG}^2 \in \mathbb{R}^{512 \times \frac{H}{8} \times \frac{W}{8} } $ from $I_{HQ}^1$ and $I_{HQ}^2$, respectively. We then employ the semantic priors $M$ to generate $N$ object masks ${ m^0, m^1,...,m^N }$, which are resized to the dimensions of $\frac{H}{8} \times \frac{W}{8}$. This allows us to obtain the mask-guided semantic features as follows:
\begin{equation}
\small
\begin{aligned}
F_{VGG}^1(n) = F_{VGG}^1\odot m^n, \
F_{VGG}^2(n) = F_{VGG}^2\odot m^n, \
\end{aligned}
\end{equation}
where $\odot$ denotes the dot product. $F_{VGG}^1(n)$ and $F_{VGG}^2(n)$ represent the semantic features guided by the $n^{th}$ mask.

Next, we calculate the semantic relationship knowledge between the mask-guided semantic features and distill this knowledge from $f^{IR2}$ to assist $f^{IR1}$ in obtaining semantic priors for improved image restoration performance. The semantic relationship is formulated as:
\begin{equation}
\small
\begin{aligned}
R_{VGG}^1(n_1,n_2) = \frac{F_{VGG}^1(n_1)F_{VGG}^1(n_2)}{\left \| F_{VGG}^1(n_1) \right \|_2\left \| F_{VGG}^1(n_2) \right \|_2 },\
\end{aligned}
\end{equation}
where $\left \| \cdot \right \|_2$ represents the L2 loss, and $R_{VGG}^1(n_1,n_2)$ represents the semantic relationship between $F_{VGG}^1(n_1)$ and $F_{VGG}^1(n_2)$. The calculation of $R_{VGG}^2(n_1,n_2)$ follows a similar approach.

To align the semantic relationships obtained from $f^{IR2}$ and $f^{IR1}$, we use the following formulation:
\begin{equation}
\small
\begin{aligned}
\mathcal{L}_{SGR} = \sum_{n_1,n_2=1,n_1\ne n_2}^{N}\frac{\left \| R_{VGG}^1(n_1,n_2)-R_{VGG}^2(n_1,n_2) \right \|_2}{N^2-N},\
\end{aligned}
\end{equation}
where $F_{SGR}$ represents the SGR loss calculated using the L2 loss function.

\subsection{Overall Optimization}
In the training stage, the IR models $f^{IR1}$ and $f^{IR2}$ undergo independent supervision using Groundtruth, with their respective reconstruction loss functions denoted as $\mathcal{L}_{recon}^1$ and $\mathcal{L}_{recon}^2$.

The IR model $f^{IR2}$ is only supervised by $\mathcal{L}_{recon}^2$ and the overall objective function of the IR model $f^{IR1}$, denoted as $\mathcal{L}$, combines these aforementioned losses as follows:
\begin{equation}
\small
\begin{aligned}
\mathcal{L} = \mathcal{L}_{recon}^1 + \lambda_1 \mathcal{L}_{SPD} + \lambda_2\mathcal{L}_{SGR},
\end{aligned}
\end{equation}
where the weights $\lambda_1$ and $\lambda_2$ are used to balance the contribution of the corresponding loss terms. Although all loss functions are optimized simultaneously, $\mathcal{L}_{SPD}$ and $\mathcal{L}_{SGR}$ only affect the gradient backpropagation of $f^{IR1}$ and are not propagated through $f^{IR2}$ with the stop-gradient mechanism. The parameters of SAM and VGG models are also frozen during the training stage.

\section{Experiments}

\begin{table*}[!t]\fontsize{8}{11}\selectfont
\begin{center}
\renewcommand\tabcolsep{5pt}
\begin{threeparttable}
\begin{tabular}{l|ccc|ccc|ccc|ccc}
\toprule[1.2pt]
\multirow{2}{*}{Method} & \multicolumn{3}{c|}{R200L} & \multicolumn{3}{c|}{R200H} & \multicolumn{3}{c|}{DID} & \multicolumn{3}{c}{DDN} \\
\cline{2-13}
& PSNR $\uparrow$& SSIM $\uparrow$& FID $\downarrow$& PSNR $\uparrow$& SSIM $\uparrow$& FID $\downarrow$& PSNR $\uparrow$& SSIM $\uparrow$& FID $\downarrow$ & PSNR $\uparrow$& SSIM $\uparrow$& FID $\downarrow$   \\
\midrule
RESCAN~\cite{li2018recurrent}&36.09 &0.9697 &- &26.75 &0.8353 &- &33.38 &0.9417 &- &31.94 &0.9345 &-       \\
PReNet~\cite{ren2019progressive}&37.80 &0.9814 &- &29.04 &0.8991 &- &33.17 &0.9481 &- &32.60 &0.9459 &-       \\
MSPFN~\cite{jiang2020multi}&38.85 &0.9827 &- &29.36 &0.9034 &- &33.72 &0.9550 &- &32.99 &0.9333 &-       \\
MPRNet~\cite{zamir2021multi}&39.47 &0.9825 &- &30.67 &0.9110 &- &33.99 &0.9590 &- &33.10 &0.9347 &-       \\
\midrule
\midrule
RCDNet~\cite{wang2020model}&39.04 &0.9846 &5.42 &30.27 &0.9063 &31.75 &34.12 &0.9561 &26.02 &33.07 &0.9483 &20.09       \\
\midrule
\multirow{2}{*}{RCDNet+ }&40.26 &0.9874 &4.19 &30.85 &0.9142 &30.05 &34.67 &0.9609  &24.00&33.84 &0.9547&19.16      \\
&  \textcolor{red}{+1.22}&\textcolor{red}{+0.0028}&\textcolor{red}{-1.23}&  \textcolor{red}{+0.58}&\textcolor{red}{+0.0079}&\textcolor{red}{-1.70}&  \textcolor{red}{+0.55}&\textcolor{red}{+0.0048}&\textcolor{red}{-2.02} &  \textcolor{red}{+0.77}&\textcolor{red}{+0.0064}&\textcolor{red}{-0.93}\\
\midrule
\midrule
Efficientderain~\cite{guo2021efficientderain} &34.42 &0.9641 &14.89 &24.20 &0.8100 &86.23 &31.99 &0.9120 &25.09 &31.75 &0.9234 &24.46       \\
\midrule
\multirow{2}{*}{Efficientderain+}&35.70 &0.9718 &9.67 &25.31 &0.8479 &56.49 &32.72 &0.9181 &21.74 &32.48 &0.9323 &20.77      \\
&  \textcolor{red}{+1.28}&\textcolor{red}{+0.0077}&\textcolor{red}{-5.22}&  \textcolor{red}{+1.11}&\textcolor{red}{+0.0379}&\textcolor{red}{-29.74}&  \textcolor{red}{+0.73}&\textcolor{red}{+0.0061}&\textcolor{red}{-3.35} &  \textcolor{red}{+0.73}&\textcolor{red}{+0.0008}&\textcolor{red}{-3.69}\\
\bottomrule[1.2pt]
\end{tabular}
\end{threeparttable}
\end{center}
\vspace{-0.6cm}
\caption{Quantitative comparison on multiple deraining datasets to evaluate our framework for the draining task. `+' represents the IR models enhanced by our proposed framework.}
\vspace{-0.4cm}
\label{derain1}
\end{table*}

\begin{figure*}[!t]
  \centering
  \includegraphics[width=0.98\textwidth]{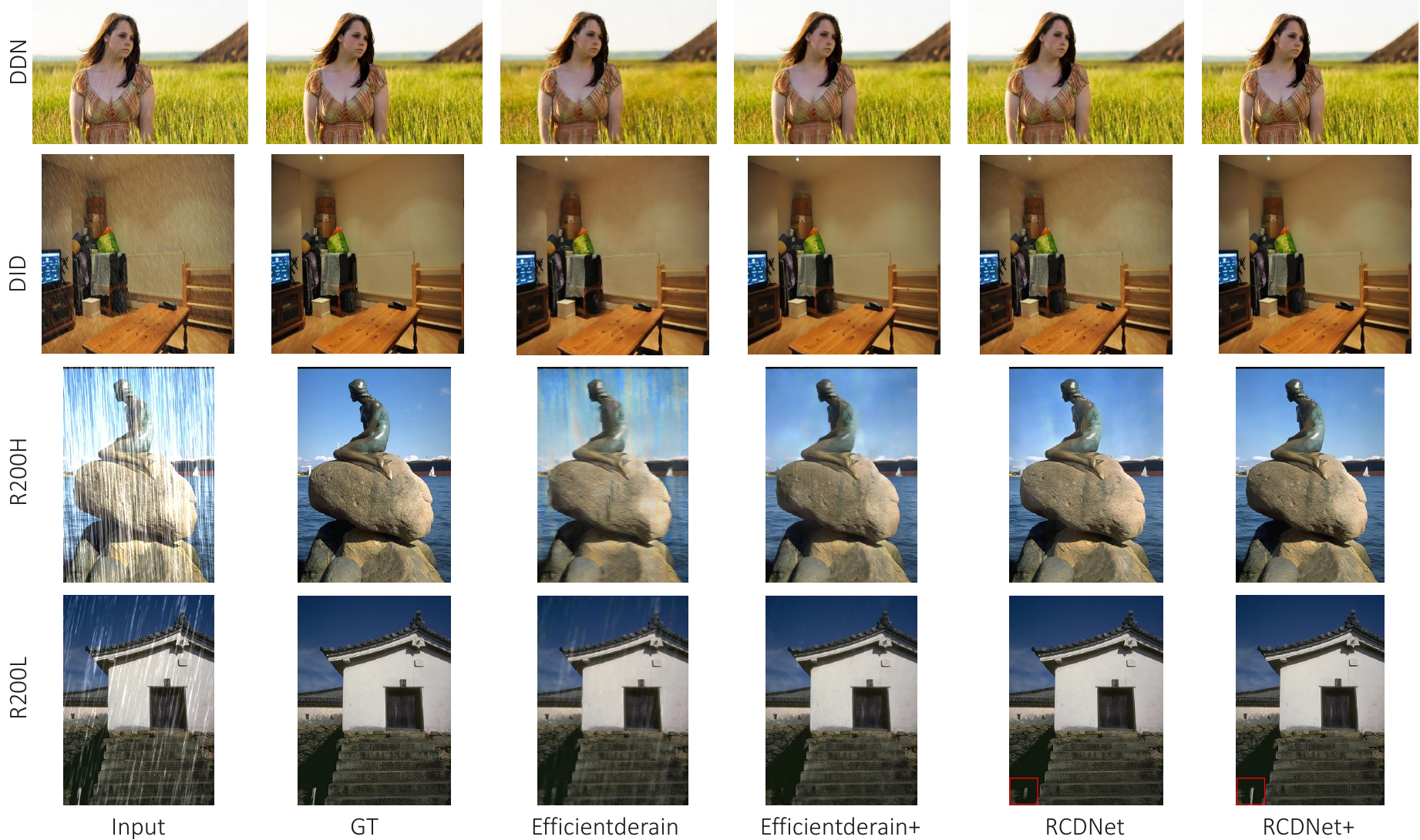}
   \vspace{-0.2cm}
  \caption{The qualitative comparison of IR models with and without our framework on various deraining datasets. } 
   \vspace{-0.6cm}
  \label{vis_derain1}
\end{figure*}

\begin{table*}[!t]\fontsize{8}{11}\selectfont
\begin{center}
\renewcommand\tabcolsep{5pt}
\begin{threeparttable}
\begin{tabular}{l|cccccc|cccccc}
\toprule[1.2pt]
\multirow{2}{*}{Method} & \multicolumn{6}{c|}{Cityscape-syn 100 mm} & \multicolumn{6}{c}{Cityscape-syn 200 mm} \\ \cmidrule(r){2-13} 
& PSNR $\uparrow$& SSIM $\uparrow$& FID $\downarrow$  & IoU $\uparrow$  & PA $\uparrow$  & DICE $\uparrow$  & PSNR $\uparrow$& SSIM $\uparrow$& FID $\downarrow$  & IoU $\uparrow$  & PA $\uparrow$  & DICE $\uparrow$ \\ \midrule
No Rain  &INF &1  &0  &0.8020  &0.9650  &0.9323   &INF &1 &0  &0.8020  &.9650  &0.9323      \\
Rain  &22.06 &0.7513  &164.66  &0.5023  &0.7452  &0.6138   &16.77 &0.6076 &253.03  &0.2128  &0.4481  &0.3076      \\
\midrule
\midrule
RCDNet~\cite{wang2020model}  &33.72 &0.9852  &9.09  &0.7869  &0.9618  &0.9263   &31.85 &0.9761 &13.69  &0.7701  &0.9585  &0.9204      \\
\midrule
\multirow{2}{*}{RCDNet+} &34.88 &0.9869   &8.37  &0.7881   &0.9620 &0.9268   &33.04  &0.9787  &12.90  &0.7741   &0.9590   &0.9214      \\
&\textcolor{red}{+1.16}& \textcolor{red}{+0.0017}& \textcolor{red}{-0.72} &\textcolor{red}{+0.0013} &\textcolor{red}{+0.0002} &\textcolor{red}{+0.0005} & \textcolor{red}{+1.19}& \textcolor{red}{+0.0026}& \textcolor{red}{-0.79} & \textcolor{red}{+0.0040} & \textcolor{red}{+0.0005} & \textcolor{red}{+0.0010}   \\
\midrule
\midrule
Efficientderain~\cite{guo2021efficientderain} &35.06   &0.9886   &8.56  &0.7807 &0.9610 &0.9250   &33.40   &0.9827 &10.71  &0.7692   &0.9592 &0.9216      \\
\midrule
\multirow{2}{*}{Efficientderain+}&36.29   &0.9912   &6.17  &   0.7862&0.9622 &0.9272   &34.65   &0.9858   &8.19  &0.7770   &0.9605 & 0.9240  \\
& \textcolor{red}{+1.23}& \textcolor{red}{+0.0026}& \textcolor{red}{-2.39} & \textcolor{red}{+0.0065} &\textcolor{red}{+0.0012} & \textcolor{red}{+0.0022}& \textcolor{red}{+1.25}& \textcolor{red}{+0.0031}& \textcolor{red}{-2.52} & \textcolor{red}{+0.0078} &\textcolor{red}{+0.0013} &\textcolor{red}{+0.0024}  \\
\bottomrule[1.2pt]
\end{tabular}
\end{threeparttable}
\end{center}
\vspace{-0.6cm}
\caption{Quantitative comparison on the synthesized deraining datasets to evaluate our framework for the downstream segmentation task.}
\vspace{-0.3cm}
\label{derain2}
\end{table*}

\begin{figure*}[!t]
  \centering
  \includegraphics[width=\textwidth]{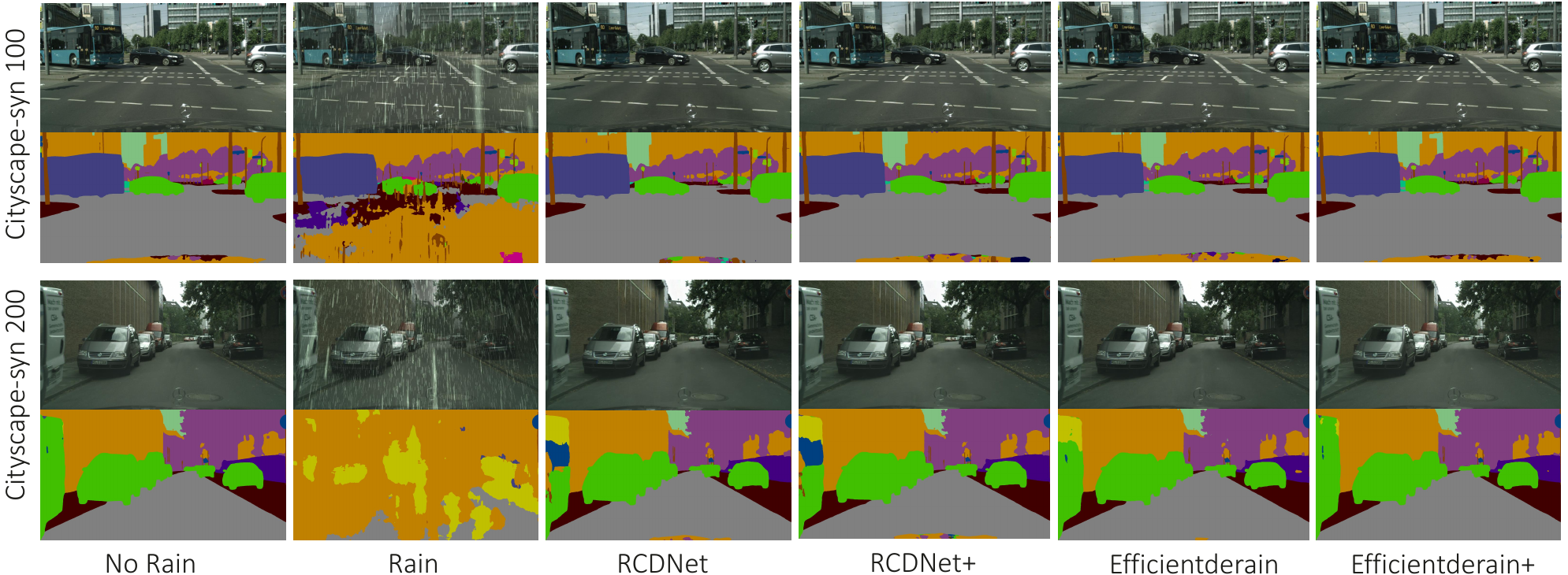}
   \vspace{-0.7cm}
  \caption{The qualitative comparison of IR models with and without our framework on the cityscape datasets. } 
   \vspace{-0.4cm}
  \label{vis_derain2}
\end{figure*}

\subsection{Datasets}

\noindent \textbf{Deraining Task.}
We evaluate our proposed framework on multiple publicly available datasets for the rain removal task, including Rain200L/H~\cite{yang2017deep}, DID~\cite{zhang2018density}, and DDN~\cite{fu2017removing}. The Rain200L/H dataset comprises 1,800 synthetic rainy images for training and 200 images for testing. The DID and DDN datasets consist of 12,000 and 12,600 synthetic images, respectively, with varying rain directions and density levels. Both the DID and DDN datasets provide 1,200 rainy images for testing. The results obtained from these datasets demonstrate the effectiveness of our method in handling diverse types of spatially varying rain streaks, and they indicate a successful reduction of rain artifacts.

Furthermore, we incorporate two additional synthetic deraining datasets, Cityscape-syn100/200~\cite{wei2022sginet}, to assess the quality of the restored images in relation to their impact on downstream segmentation tasks. These datasets are synthesized based on the Cityscape dataset~\cite{cordts2016cityscapes}, allowing us to evaluate image quality using downstream segmentation metrics on the validation set.

\noindent \textbf{Deblurring Task.}
We conduct an evaluation of our framework on the GoPro dataset~\cite{nah2017deep} for the image deblurring task. The GoPro dataset contains a total of 2,103 training images and 1,111 test images. To generate the blurs of different strengths, a varying number of successive latent frames are averaged together. The images in this dataset are captured using a GoPro camera at a frame rate of 240 fps.

\noindent \textbf{Denoising Task.}
We conduct an evaluation of our framework on the SenseNoise dataset~\cite{zhang2022idr} for the image denoising task. The SenseNoise dataset comprises 500 diverse scenes, each consisting of high-resolution images. The dataset includes both indoor and outdoor scenes, and it provides high-quality ground truth images for reference.

\subsection{Evaluation Metrics}

We employ two commonly used metrics, Peak Signal-to-Noise Ratio (PSNR)~\cite{huynh2008scope} and Structural Similarity Index (SSIM)~\cite{wang2004image}, to evaluate the performance of our image restoration tasks. Additionally, we introduce the Fréchet Inception Distance (FID)~\cite{heusel2017gans} as a measure of the subjective visual quality perceived by humans. In the context of the cityscape-syn datasets, we utilize pixel accuracy (PA), intersection over union (IOU), and DICE~\cite{shamir2019continuous} as segmentation metrics to assess the performance of the downstream segmentation tasks.

\subsection{Implementation Details}
All the components of our framework are trained simultaneously. We utilize the PyTorch platform within the Python environment and employ NVIDIA Tesla V100 GPUs with 32 GB memory for training. The framework is trained using the Adam optimizer with $\beta_1 = 0.9$ and $\beta_2 = 0.999$. We set the learning rate to $1e-4$ and utilize a batch size of 8 for a total of 200 epochs.

\subsection{Selected Baseline Methods}

In order to demonstrate the effectiveness of our general framework, we conduct experiments using several well-established image restoration models. For the rain removal task, we select two representative models: RCDNet~\cite{wang2020model} and Efficientderain~\cite{guo2021efficientderain}. For the deblurring and denoising tasks, we chose the widely recognized Uformer model~\cite{wang2022uformer} as our representative model. These models are carefully selected to cover a range of restoration tasks and showcase the versatility of our framework.


\subsection{Quantitative and Qualitative Results}
\noindent \textbf{Deraining Task.}
In Table~\ref{derain1} and Table~\ref{derain2}, we provide the validation results of our framework on the deraining task. The tables illustrate that our framework consistently outperforms the original RCDNet and Efficientderain models, yielding significant performance gains of both objective and subjective visual metrics across multiple datasets. Specifically, our framework achieves an average improvement of 0.91 dB PSNR over the RCDNet model and 1.02 dB PSNR over the Efficientderain model. These improvements are primarily attributed to the introduction of advanced capabilities for suppressing noise and artifacts while preserving texture and color consistency. Furthermore, we evaluate the segmentation performance of different restored images using an HRNet~\cite{sun2019high} model pre-trained on the Cityscape dataset, as shown in Table~\ref{derain2}. The results demonstrate that our framework consistently achieves downstream segmentation improvements in terms of IoU, PA, and DICE metrics and significantly better segmentation results compared to the segmentation results of degraded images in `Rain'.

In Fig.~\ref{vis_derain1}, we visualize the deraining results on the benchmark datasets. It can be seen that our framework assists the existing IR models not only in removing rain streaks more effectively but also in the preservation of high-fidelity object boundaries. 

We also visually depict the deraining and corresponding segmentation results on the Cityscape-syn datasets in Figure~\ref{vis_derain2}. From the deraining perspective, our framework effectively aids the IR models in removing a larger number of rain streaks and restoring the intricate structure and content of rainy images. Additionally, our framework provides enhanced semantic priors to the image restoration models, enabling the generation of restored images with richer semantic information. This, in turn, facilitates better segmentation results for the downstream segmentation task. The incorporation of semantic priors through our framework enhances both deraining and segmentation performance, demonstrating its effectiveness in addressing the challenges posed by rainy images.

\begin{figure}[!t]
  \centering
  \includegraphics[width=0.47\textwidth]{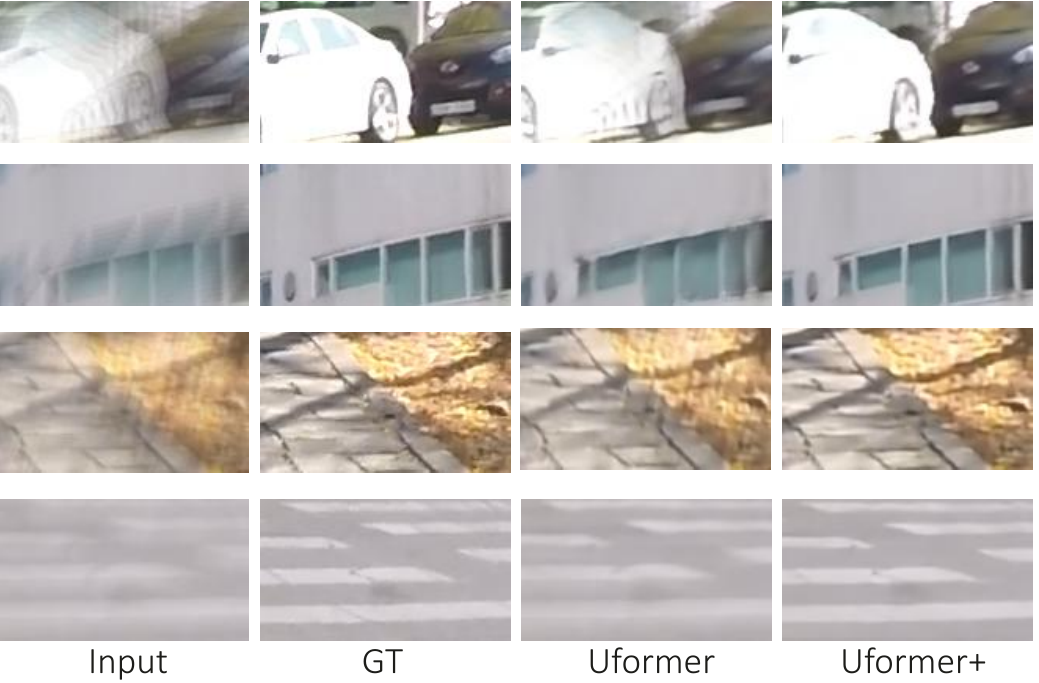}
   \vspace{-0.3cm}
  \caption{The qualitative comparison of IR models with and without our framework on the GoPro dataset. } 
   \vspace{-0.6cm}
  \label{vis_blur}
\end{figure}

\begin{figure*}[!t]
  \centering
  \includegraphics[width=\textwidth]{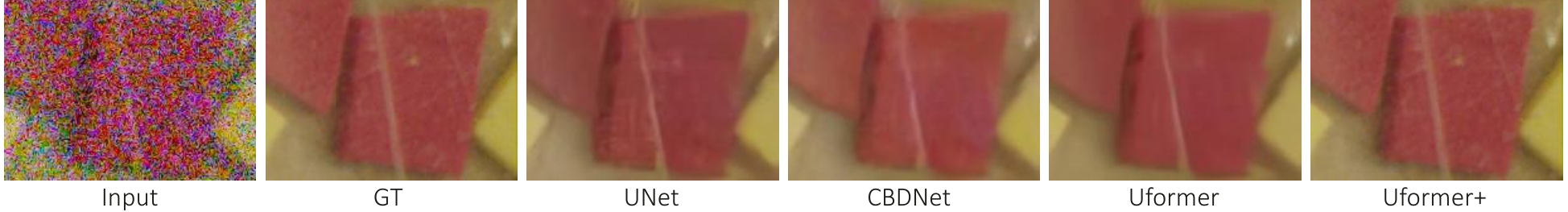}
   \vspace{-0.7cm}
  \caption{The qualitative comparison of IR models with and without our framework on the SenseNoise dataset. } 
   \vspace{-0.5cm}
  \label{vis_noise}
\end{figure*}

\noindent \textbf{Deblurring and Denoising Tasks.}
We validate our framework on the deblurring and denoising tasks In Table~\ref{deblur} and Table~\ref{denoise} with the representative model Uformer, respectively. It can be observed that our framework achieves over 0.1 dB PSNR performance improvement of Uformer without inferring with its inference processing on both of these two tasks.

%

In Figure \ref{vis_blur}, we present visual results on the GoPro dataset for the deblurring task. These visualizations clearly demonstrate the effectiveness of our framework in enhancing the performance of the Uformer model. Our framework enables the Uformer model to effectively handle dynamic blurring and significantly improve the quality of the deblurred images. Moving on to Figure \ref{vis_noise}, we showcase visual results on the SenseNoise dataset for the denoising task. These results further illustrate the effectiveness of our framework in enhancing the performance of the Uformer model. This indicates that our framework enhances the denoising capabilities of the Uformer model, resulting in improved image quality and more accurate noise reduction.



\begin{table}[!t]\fontsize{8}{11}\selectfont
\begin{center}
\renewcommand\tabcolsep{12pt}
\begin{tabular}{l|ccc}
\toprule[1.2pt]
 Method & PSNR $\uparrow$& SSIM $\uparrow$& FID $\downarrow$  \\
 \midrule
DeblurGAN~\cite{kupyn2018deblurgan} &28.70 & 0.858 &-        \\
DeblurGANv2~\cite{kupyn2019deblurgan} &29.55 &0.934 &-       \\
SRN~\cite{tao2018scale}  &30.26 &0.934 &-     \\
DBGAN~\cite{zhang2020deblurring} &31.10 &0.942 &-        \\
MT-RNN~\cite{park2020multi} &31.15 &0.945 &-       \\
DMPHN~\cite{zhang2019deep}  &31.20   &0.940 &-     \\
 \midrule
 \midrule
Uformer~\cite{wang2022uformer} &32.10   &0.949 &11.36       \\
\midrule
\multirow{2}{*}{Uformer+}  &32.21   &0.950 &11.10     \\
&\textcolor{red}{+0.11}& \textcolor{red}{+0.001}& \textcolor{red}{-0.26}      \\
\bottomrule[1.2pt]
\end{tabular}
\end{center}
\vspace{-0.5cm}
\caption{Quantitative comparison on the GoPro dataset to evaluate our framework for the deblurring task. }
\vspace{-0.4cm}
\label{deblur}
\end{table}



\begin{table}[!t]\fontsize{8}{11}\selectfont
\begin{center}
\renewcommand\tabcolsep{12pt}
\begin{tabular}{l|ccc}
\toprule[1.2pt]
 Method & PSNR $\uparrow$& SSIM $\uparrow$& FID $\downarrow$  \\
 \midrule
UNet~\cite{ronneberger2015u} &34.92   &0.9130 &20.61     \\
CBDNet~\cite{guo2019toward} &35.00   &0.9140 &17.73     \\
 \midrule
 \midrule
Uformer~\cite{wang2022uformer} &35.14 &0.9139 &17.37       \\
\midrule
\multirow{2}{*}{Uformer+}  &35.25   &0.9151 &17.09     \\
& \textcolor{red}{+0.11} & \textcolor{red}{+0.0012} & \textcolor{red}{-0.28}    \\
\bottomrule[1.2pt]
\end{tabular}
\end{center}
\vspace{-0.5cm}
\caption{Quantitative comparison on the SenseNoise dataset to evaluate our framework for the denoising task. }
\vspace{-0.5cm}
\label{denoise}
\end{table}


\subsection{Ablation Study}

\noindent \textbf{Components of the Proposed Framework.}
In Table~\ref{ablation}, we present the results of an ablation study conducted on the main components of our proposed framework. The cascaded networks $f^{IR1}$ and $f^{IR2}$ in our framework demonstrate improved performance, as the cascading architecture enhances the output of $f^{IR2}$. Furthermore, by incorporating the semantic priors from SAM using the SPF scheme, we observe further performance improvement. Both the SPD scheme and the SGR module consistently enhance the performance of the output of $f^{IR1}$ while maintaining the performance of $f^{IR2}$ itself. Notably, we achieve comparable performance between the outputs of $f^{IR1}$ and $f^{IR2}$.

In addition, we evaluate the impact of different semantic priors obtained from SAM and the commonly used segmentation model, PSPNet. As SAM provides more detailed and comprehensive semantic priors through segmentation at various levels of granularity, our framework, which integrates SAM, achieves superior performance compared to incorporating semantic priors from instance and category labels alone.

\begin{table}[!t]\fontsize{7.5}{12}\selectfont
\begin{center}
\renewcommand\tabcolsep{2.5pt}
\begin{threeparttable}
\begin{tabular}{l|ccc|ccc}
\toprule[1.2pt]
\multirow{2}{*}{Method}& \multicolumn{3}{c|}{$f^{IR1}$} & \multicolumn{3}{c}{$f^{IR2}$} \\ \cline{2-7} 
& PSNR $\uparrow$& SSIM $\uparrow$& FID $\downarrow$& PSNR $\uparrow$& SSIM $\uparrow$& FID $\downarrow$    \\ \midrule
$f^{IR1}$  &34.09  &0.9847 &5.02 &- &- &-        \\
+ $f^{IR2}$&34.01 &0.9831 &5.13 &34.21 &0.9851 &4.68        \\
+ SPF, (w. SAM)&34.03  &0.9836 &5.11 &34.89 &0.9859 &4.67        \\
 + SPD, (w. SAM)&34.84  &0.9853 &4.71 &34.91 &0.9865 &4.52        \\
+ SGR, (w. SAM) &\textbf{35.29}  &\textbf{0.9864} &\textbf{4.49} &\textbf{35.30} &\textbf{0.9864} &\textbf{4.51}        \\
\midrule
 + SGR, (w. PSPNet) &35.08  &0.9858 &4.62 &35.09 &0.9858 &4.65      \\
\bottomrule[1.2pt]
\end{tabular}
\end{threeparttable}
\end{center}
\vspace{-0.5cm}
\caption{Albaltion studies on components of our framework.}
\vspace{-0.3cm}
\label{ablation}
\end{table}

\begin{table}[!t]\fontsize{7.5}{9}\selectfont 
    \begin{center}
    \renewcommand\tabcolsep{7pt}
    \begin{threeparttable}
    \begin{tabular}{c|ccc}
    \toprule[1.2pt]
          
    $\lambda_1$ & 0.0005 & 0.005 & 0.05  \\
    \midrule 
    PSNR / SSIM   &35.12 / 0.9859   &\textbf{35.29} / \textbf{0.9864} & 35.16 / 0.9861  \\
    \midrule
    \midrule
    $\lambda_2$ & 20 & 200 & 2000  \\
    \midrule
    PSNR / SSIM  &35.26 / 0.9863   &\textbf{35.29} / \textbf{0.9864}  & 35.21 / 0.9863 \\
    \bottomrule[1.2pt]
    \end{tabular} 
    \end{threeparttable}
    \end{center}
 \vspace{-0.5cm}
 \caption{Ablation study on loss weights of our framework.}
 \label{hyper}
 \vspace{-0.5cm}
\end{table}

\noindent \textbf{Hyperparameters.}
We conduct ablation experiments to investigate the impact of hyperparameters $\lambda_1$ and $\lambda_2$ in balancing the losses $\mathcal{L}_{PSD}$ and $\mathcal{L}_{SGR}$, respectively. The results of these experiments are summarized in Table~\ref{hyper}. Various values are tested to identify an optimal weight for each hyperparameter. Based on the experimental findings, we empirically set $\lambda_1 = 0.005$ and $\lambda_2 = 200$ as they yield the best performance. These values are chosen to strike a balance between the two losses and achieve optimal results for our framework.

\subsection{Comparison of existing SAM priors introduction methods}

To thoroughly validate the superiority of our proposed framework over existing methods that incorporate SAM's priors to enhance the image deblurring task, we conducted further testing for the advanced image restoration (IR) model, NAFNet~\cite{chen2022simple}, on two widely used datasets, including the GoPro~\cite{nah2017deep} dataset and the ReLoBlur~\cite{li2023real} dataset.

In our experiments, we employ the NAFNet architecture with 32 channels as the image deblurring model. We compare three different training methods: one utilizing the concatenation method (CAT) for incorporating SAM priors as proposed in~\cite{jin2023let}, the second method exploiting only the MAP Unit of SAM-Deblur~\cite{li2023sam}, and finally, the SAM-Deblur framework, which combines both the MAP Unit and mask dropout. SAM-Deblur currently represents the state-of-the-art method that leverages SAM's priors to enhance image restoration tasks.

As illustrated in Table~\ref{comparison}, the performance of the aforementioned methods on the datasets mentioned clearly showcases the effectiveness and superiority of our proposed framework compared to existing approaches that rely on SAM's priors. It is worth noting that our framework not only achieves better performance but also preserves the inference efficiency of the original IR models without the need for SAM in the inference stage.

\begin{table}[!h]\fontsize{8}{11}\selectfont
\begin{center}
\renewcommand\tabcolsep{8pt}
\begin{threeparttable}
\begin{tabular}{l|cc|cc}
\toprule[1.2pt]
\multirow{2}{*}{Method} & \multicolumn{2}{c|}{GoPro}  & \multicolumn{2}{c}{ReLoBLur} \\
\cline{2-5}
& PSNR $\uparrow$& SSIM $\uparrow$ & PSNR $\uparrow$& SSIM $\uparrow$   \\
\midrule
NAFNet~\cite{chen2022simple}&32.85 &0.960   &25.26 &0.687        \\
 + CAT~\cite{jin2023let}&32.88 &0.961    &29.77 &0.882        \\
 + MAP~\cite{li2023sam}&32.82 &0.960    &30.86 &0.897        \\
 + SAM-Deblur~\cite{li2023sam}&32.83 &0.960  &32.29 &0.903        \\
 + Ours &\textbf{32.90}  &\textbf{0.961}   &\textbf{32.35}&\textbf{0.904}         \\
\bottomrule[1.2pt]
\end{tabular}
\end{threeparttable}
\end{center}
\caption{Quantitative comparison of different methods that introduce SAM priors to boost the deblurring task. }
\label{comparison}
\end{table}

\section{Conclusion}
We propose a general framework to distill the semantic knowledge of the segment anything model (SAM) and boost existing image restoration (IR) models. By incorporating the semantic priors fusion (SPF) and semantic priors distillation (SPD) schemes, we successfully enhance the performance of multiple IR models across tasks such as deraining, deblurring, and denoising. Our framework addresses the computational cost limitations of SAM while effectively leveraging its semantic priors.

\vspace{-0.1cm}
\section*{Acknowledgement}
\vspace{-0.1cm}
This work was supported by the National Key R\&D Program of China (2022YFB4701400/4701402), in part by the SSTIC Grant (KJZD20230923115106012), in part by Shenzhen Key Laboratory (ZDSYS20210623092001004), and in part by the Beijing Key Lab of Networked Multimedia. We gratefully acknowledge the support of MindSpore, CANN (Compute Architecture for Neural Networks) and Ascend AI Processor used for this research.

\begin{figure*}[!h]
  \centering
  \includegraphics[width=\textwidth]{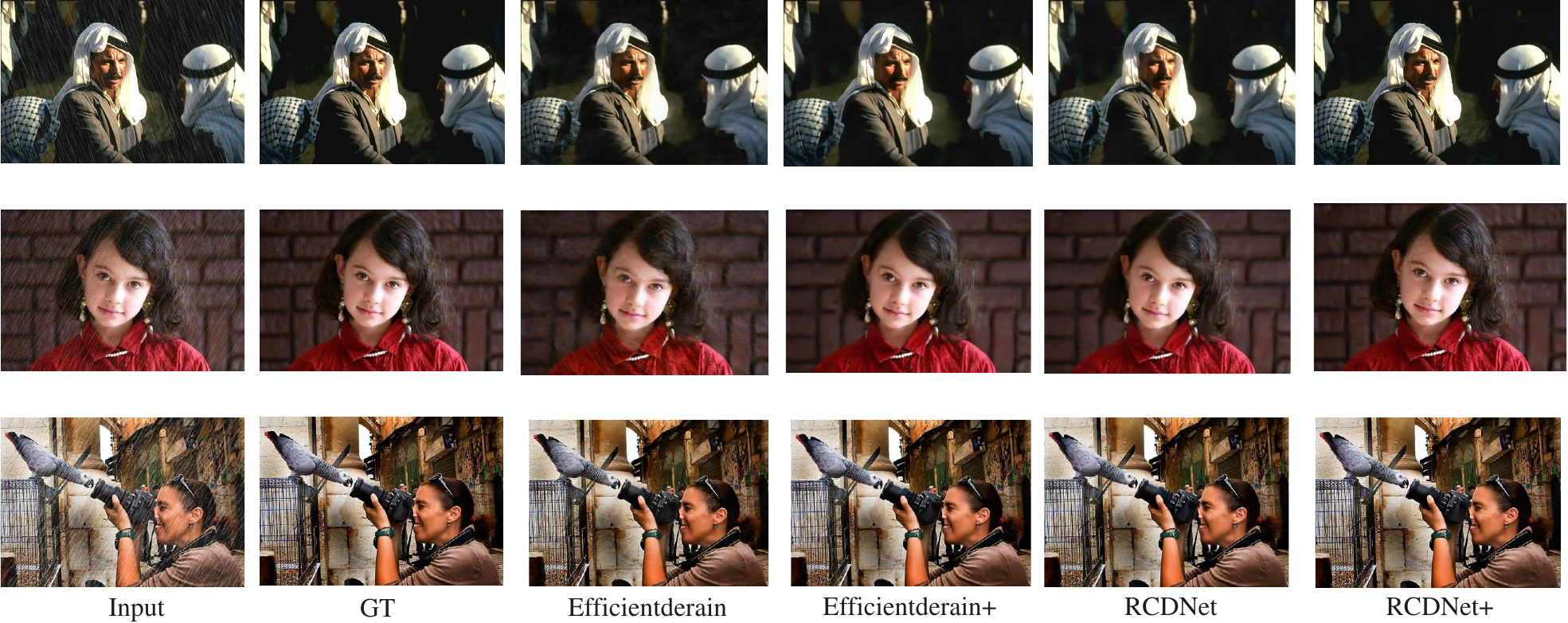}
  \caption{The qualitative comparison of IR models with and without our framework on the DDN dataset for the deraining task. } 
  \label{ddn_crop}
\end{figure*}

\begin{figure*}[!h]
  \centering
  \includegraphics[width=\textwidth]{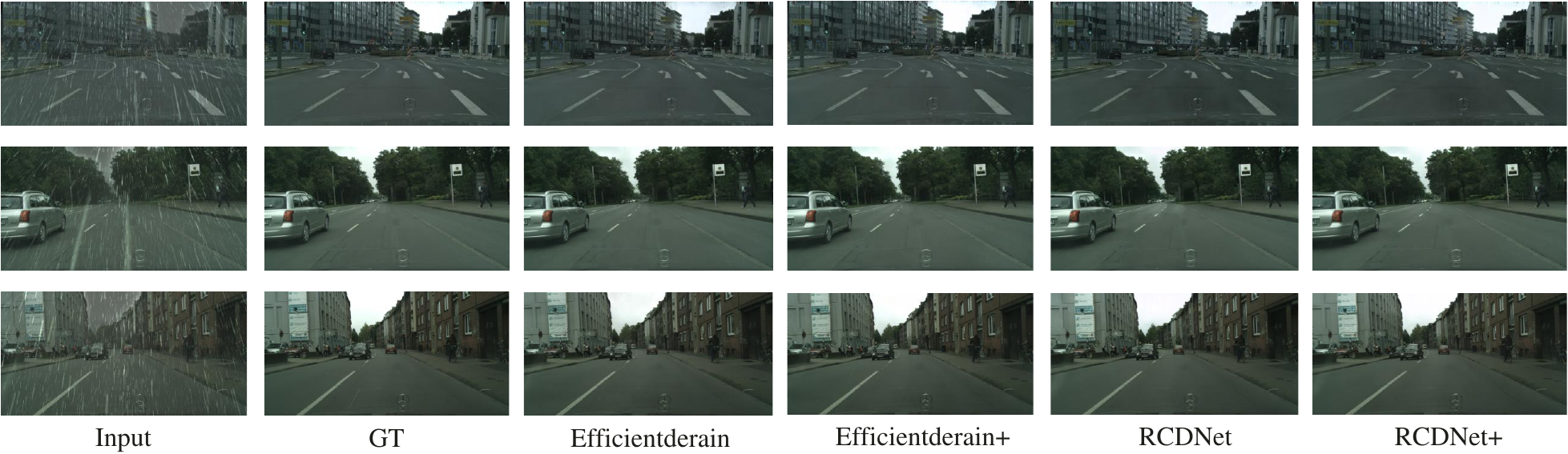}
  \caption{The qualitative comparison of IR models with and without our framework on the cityscape-syn 100 dataset for the deraining task. } 
  \label{city100_crop}
\end{figure*}

\begin{figure*}[!h]
  \centering
  \includegraphics[width=\textwidth]{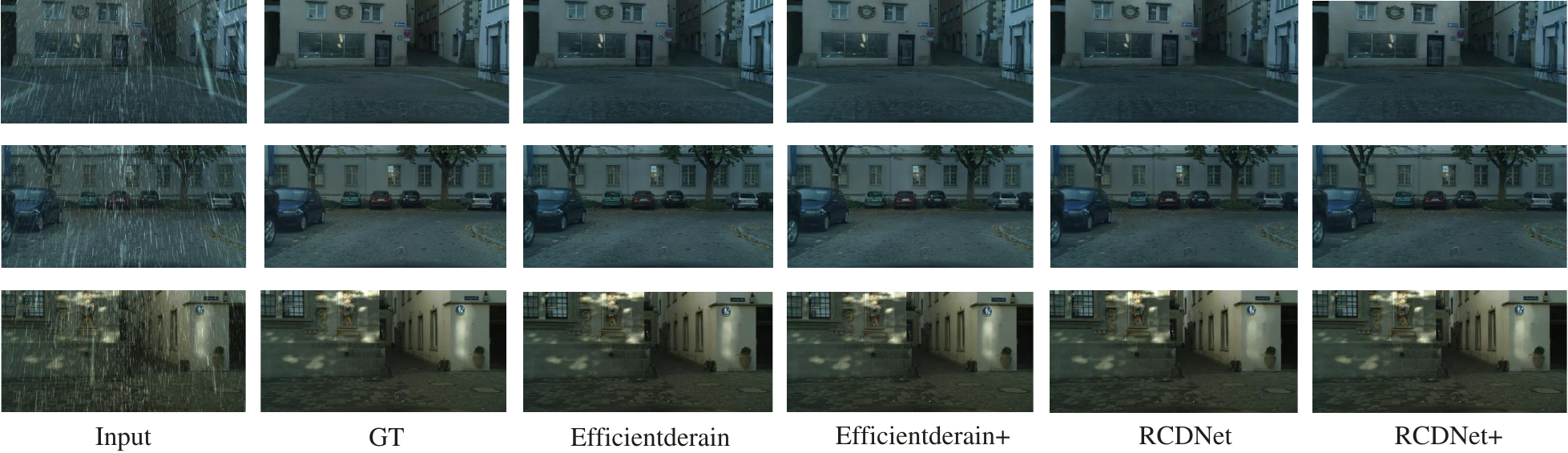}
  \caption{The qualitative comparison of IR models with and without our framework on the cityscape-syn 200 dataset for the deraining task. } 
  \label{city200_crop}
\end{figure*}

\begin{figure*}[!h]
  \centering
  \includegraphics[width=\textwidth]{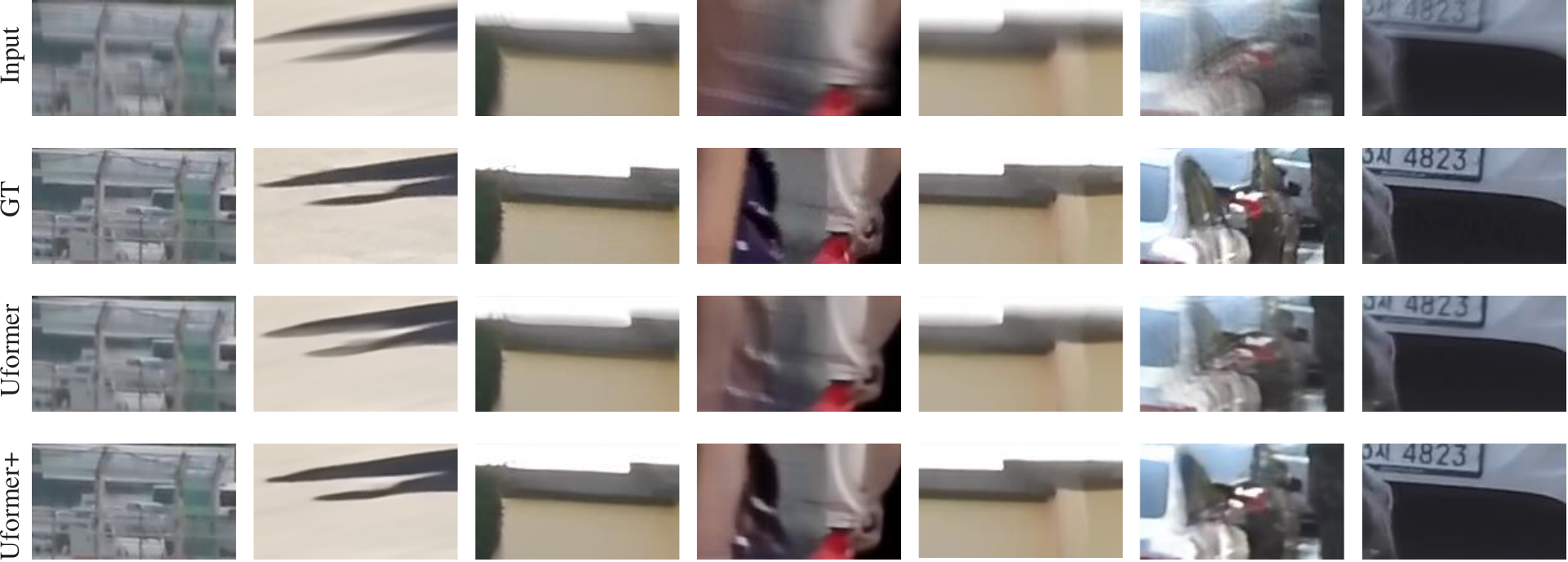}
  \caption{The qualitative comparison of IR models with and without our framework on the GoPro dataset for the deblurring task. } 
  \label{city200_crop}
\end{figure*}

\clearpage
\newpage

{\small
\bibliographystyle{ieee_fullname}
\bibliography{egbib}
}

\end{document}